\documentclass{article}

\usepackage{microtype}
\usepackage{graphicx}
\usepackage{subfigure}
\usepackage{amssymb,amsthm,amsmath,latexsym}
\usepackage{booktabs} 

\usepackage{hyperref}

\usepackage[accepted]{icml2018}

\icmltitlerunning{Playing against Nature: causal discovery for decision making under uncertainty}

\begin{document}

\twocolumn[
\icmltitle{Playing against Nature: causal discovery for decision making under uncertainty}



\icmlsetsymbol{equal}{*}

\begin{icmlauthorlist}
\icmlauthor{Mauricio Gonzalez-Soto}{inaoe}
\icmlauthor{Luis Enrique Sucar}{inaoe}
\icmlauthor{Hugo Jair Escalante}{inaoe}

\end{icmlauthorlist}

\icmlaffiliation{inaoe}{Department of Computer Science, National Institute of Astrophysics Optics and Electronics (INAOE), Mexico}

\icmlcorrespondingauthor{Mauricio Gonzalez-Soto}{mauricio@inaoep.mx}
\icmlcorrespondingauthor{Luis E. Sucar}{esucar@inaoep.mx}
\icmlcorrespondingauthor{Hugo J. Escalante}{hugojair@inaoep.mx}

\icmlkeywords{Causality, Causal Inference, Game Theory, Decision under Uncertainty}

\vskip 0.3in
]



\printAffiliationsAndNotice{\icmlEqualContribution} 

\begin{abstract}
We consider decision problems under uncertainty where the options available to a decision maker and the resulting outcome are related through a causal mechanism which is unknown to the decision maker. We ask how a decision maker can learn about this causal mechanism through sequential decision making as well as using current causal knowledge inside each round in order to make better choices had she not considered causal knowledge and propose a decision making procedure in which an agent holds \textit{beliefs} about her environment which are used to make a choice and are updated using the observed outcome. As proof of concept, we present an implementation of this causal decision making model and apply it in a simple scenario. We show that the model achieves a performance similar to the classic Q-learning while it also acquires a causal model of the environment.
\end{abstract}

\section{Introduction}
A fundamental part of intelligent reasoning is being able to make decisions under uncertain conditions (\cite{danks2014unifying}, \cite{lake2017building}, \cite{pearlwhy}).
In some cases, a decision maker who faces an uncertain environment has enough information to make choices by maximizing expected utility, which is the classic formal criteria for making decisions if rational preferences are assumed (\cite{bernardo2000bayesian}, \cite{gilboa2009decision}). On the other hand, if enough information is not available, the decision maker could attempt to \textit{learn} from the environment by interacting with it.

Learning by interaction has been extensively studied by computer scientists using the Reinforcement Learning (RL) setting \cite{sutton1998reinforcement}, but the most common used techniques  in this field are purely associative and do not consider any high-level structure of the environment beyond what is expressable in a Markov Decision Process \cite{garnelo2016towards}.

A particular case of a \textit{higher level structure}; i.e., beyond associative patterns, is the case of \textit{causal} structure. A causal structure encondes a series of \textit{cause-effect} relations between events and knowing such relations allows a decision maker to add extra knowledge into the uncertainty of his environment and also allows it to plan ahead his actions since he can predict what a certain action will cause (\cite{spirtes2000causation}, \cite{pearl2018theoretical}).

Since human beings are known to learn causal models in sequential decision making processes (\cite{sloman2006causal}, \cite{nichols2007decision}, \cite{meder2010observing}, \cite{hagmayer2013repeated}, \cite{danks2014unifying}), and even though this learning is not perfect \cite{rottman2014reasoning}, we propose that an autonomous agent can learn and use causal information while interacting with an uncertain environment which is governed by a fixed \textit{causal mechanism} which is unknown to the agent.  

While the standard setting in RL is to model the agent-environment interaction as an agent that moves from one \textit{state} to another inside a model of the environment and observing a reward as these transitions occur, we propose to model it as a \textit{game} between the decision maker and a player called \textit{Nature} which will select his actions from the causal model in response to what the decision maker has chosen. 

The proposed way for an agent to learn from repeated interactions is by giving her \textit{beliefs} about the structure of the environment and a way to update them after an outcome has been observed. The agent, using her current beliefs, will generate a \textit{local} causal model and choose an action from it as if that model was the true one. Then, after she observes the consequences of her actions, her beliefs will be updated according to the observed information in order to make a better choice the next time. The agent, besides learning a policy to choose actions will also learn a causal model from the environment since the causal model she forms will approximate the true model.

Learning a causal model of the environment allows to extract high-level insights of a phenomena beyond associative descriptions of what is observed. A causal model is able to \textit{explain} why a particular decision was made since it allows to extract the causes and effects of an agent's actions. Once a causal model is acquired, an external user is able to reason about \textit{what...if...} statements that associative methods can not answer \cite{pearl2018why}. When a decision maker chooses an action out of many, a causal model allows to ask what would've happened if another action was taken without actually performing the alternative action.

\section{Related Work}
Decision problems in which the actions available to a decision maker are interventions over a known causal model are analyzed by \cite{lattimoreNIPS2016} as a \textit{bandit problem} where the optimal action must be learned over $T$ rounds of action-observation in which only one action can be chosen. In a classic bandit problem an agent chooses an \textit{arm} from a slot machine, observes a reward and then moves on to the next machine which is of the same kind and whose initial settings are independent of the previous machine and action \cite{sutton1998reinforcement}.

Several algorithms exists for finding the best arm in a multi-arm bandit, such as those described in \cite{bubeck2009pure}, \cite{audibert2010best}, \cite{gabillon2012best}, \cite{agarwal2014taming} , \cite{jamieson2014lil},  \cite{jamieson2014best}, \cite{chen2015optimal}, \cite{carpentier2016tight}, \cite{russo2016simple},  \cite{kaufmann2016complexity}, but none of these works consider causal-governed environments. In \cite{ortega2014generalized} results on sequential decision-making using Generalized Thompson Sampling that could be extended into causal inference problems are given.

As far as we know \cite{lattimoreNIPS2016} is the first paper to consider causal relations between the effects of actions. They consider a decision maker who must choose the best among several possible interventions on a given causal model. The optimality of the action in this context is in terms of the minimal regret. The case where the causal model is not known is left as future work.

By considering a causal model which is partially known and intervening variables from the unknown part of the model and by avoiding sampling arms that are considered sub-optimal, \cite{sen2017identifying} extend the work of \cite{lattimoreNIPS2016}.

The aforementioned papers assume the causal model is known to the decision makers so their work focuses on \textit{using} causal information to make good choices, but the problem of \textit{acquiring} this causal knowledge is left unattacked.

In this work we propose to acquire, by repeated interaction, causal information about the environment as well as using the current causal knowledge inside each round to make better decisions. By modelling the environment as a player in a game we allow it to have objectives to pursue which will allow to model a rich family of situations where several agents are competing against each other and a \textit{causal entity} controls the outcomes. 

\section{Problem setup}{\label{problem_setup}}
By \textit{causality} we mean a stochastic binary relation between events of a probability space $(\Omega, \mathcal{F}, \mathbb{P})$ denoted by $“\to”$ that is transitive, irreflexive and antisymmetric \cite{spirtes2000causation}. 

A directed acyclic graph (DAG) can be used to represent all of the relations that occur in that space by considering a node for every variable that is related to another and a directed edge to express the causal relation, call this DAG $\mathcal{G}$ and consider a probability measure $P_{\mathcal{G}}$ that expresses the conditional statements from the DAG. 

We require that this measure satisfies the Markov Causal Condition, Causal Minimality and Causal Faithfulness as stated in \cite{spirtes2000causation}. The relation between $\mathbb{P}$ and $P_{\mathcal{G}}$ is given by the Manipulation Theorem of \cite{spirtes2000causation} and the Do-Calculus rules from \cite{pearl2009causality}. We also require that the condition known as \textit{Causal Sufficiency} is satisfied by the model, which means there are not any causes lying \textit{outside} of the model. 

Let $(\mathcal{A},\mathcal{E},\mathcal{C},\succeq)$ a Decision Problem under Uncertainty in which an agent has to choose one among several options $a \in \mathcal{A}$ which are causally related to the elements of $\mathcal{C}$. The elements in $\mathcal{E}$ are uncertain events which are governed by a causal mechanism. This means that when the decision maker chooses an action $a$, an uncertain event will occurr in such a way that an outcome $c$ will occur, which is causally related to $a$. The decision maker seeks to maximize her utility and we assume that she has \textit{rational preferences}, so we can substitute her preferences $\succeq$ for the expected value of a utility function $u$ \cite{gilboa2009decision}. If the decision maker does not know the probabilities nor the structure of the underlying causal model then she can not calculate the expected utility of any action. Instead, she will have a \textit{subjective} probability distribution which represents the agent's knowledge and uncertainties which will be updated by interacting with the environment through succesive rounds of decision making. 

Inside each round, any response from the environment will be independent from the previous rounds, but the actions of the decision maker will be based upon previously acquired causal information and are expected to improve the utility for the agent.

We define a game between the decision maker and a new abstract player called Nature. The base game's structure will be the same as the original decision problem. Nature will be indifferent among the possible outcomes of the game and will select its actions from the causal model. This is interpreted as the \textit{causal response} from the environment to the actions of the decision maker. Nature having objectives to pursue (non-constant payments) will be left as future work.

For an agent to reason about and modify her causal knowledge we endow her with a probability distribution $p(\theta)$ over a suitable space. The beliefs must allow to form a \textit{local} model in a given moment to be used for decision making. We will later exploit the fact that causal graphical models can be expressed in terms of conditional distributions, so having beliefs about a causal model is equivalent to having beliefs about these conditional distributions. After each round of the game, the beliefs will be updated in a Bayesian way in order to achieve convergence towards the true model \cite{shoham2008multiagent}.

\section{Proposed Method}
In this section we describe our approach for studying decision making in causal environments as described in Section \ref{problem_setup}. For the sake of explanation we consider three separate cases:

\begin{itemize}
\item The decision maker fully knows the causal model.
\item The decision maker knows only the structure of the causal model.
\item The decision maker does not know the causal model.
\end{itemize}

\subsection{The causal model is completely known}{\label{known}}
Consider a decision problem under uncertainty where a decision maker has to choose on out of many elements of a set $A$ and where the consequences, or effects, of her actions are expressed as the outcome of a random variable $Y$ which we will call \textit{target variable}. The relation between values of $Y$ and actions $a \in A$ is expressed by a causal graphical model $\mathcal{G}$, which is known by the decision maker. The decision maker whishes to choose an action $a \in A$ such that the observed value of $Y$ maximizes her utility. It is assumed that the variable that is going to be intervened is known by the decision maker; i.e., she knows what variable can she intervene.

This is the simplest case of the three mentioned because if the decision maker fully knows the causal model, then she can proceed as in classic decision problems by directly obtaining the probabilities of different values for the target variable given that an action is made and choose $a$ which achieves the highest probability for the desired value of the target variable. The action selected will be a \textit{best response} for the decision maker as well as the maximum expected utility choice.

Pearl's do-calculus \cite{pearl2009causality} says that the effect of setting some variable $X_i$ to a value $x_j$ can be expressed in terms of \textit{observational} distributions as follows:
\[ P(X_1,...,X_n | do(X_i = j )) = \prod_{k \neq i} P(X_k | \textrm{Pa}(X_k)). \]

The decision maker can use this expression to find the probabilities for her desired value of the target variable given the possible interventions available to her.

\subsection{Only the structure is known}
If only the graphical structure of $\mathcal{G}$ is known, then it is not obvious how to find the best action to make since the information required for calculating expected utilities is not available.

In this case, the decision maker will attempt to learn from her uncertain environment by forming \textit{beliefs} over unknown parameters of the environment and update them according to the observed outcomes. In order to make a Bayesian update over the parameters, these must be defined in such way that in each round the decision maker can define a causal model from the parameters in order to make a decision using this model as if it was the true one as described in Section \ref{known}.

To model the interaction between the decision maker and the environment, we consider a game with the following characteristics:
\begin{itemize}
\item \textbf{Players:} The set of players of this game is the set whose elements are the original decision maker, and a player called Nature.
\item \textbf{Actions:} The actions for the decision maker are the available options she has in the decision problem;i.e. $A$. The actions of Nature are possible realizations of the variables of the causal model. 
\item \textbf{Preferences:} The decision maker satisfies the von Neumann-Morgenstern axioms of rationality and therefore it is assumed to be maximizing expected utilities. Nature is indifferent over outcomes.
\item \textbf{Beliefs:} Since the decision maker has uncertainty about her environment, she will encode it in a probability distribution $p(\theta)$ over a suitable space.
\end{itemize}

In this game we will assume that Nature moves first and assigns some \textit{state} to the environment which is unknown to the decision maker. For this reason, the base game is an extensive game with imperfect information since the decision maker makes a choice without knowing the play made by Nature. We choose extensive games since Nature's moves are interpreted as a \textit{response} from the causal model to the actions of the decision maker, so a sequential interpretation had to be considered.

Since the decision maker knows the graph structure, she can explicitly find a non interventional expression for the interventional distribution and update her beliefs about these unknown quantities from observed data. If the decision maker were not allowed to know, at the end of each round, the play of the Nature then this will have to be estimated as a hidden variable using, for example, the EM algorithm \cite{dempster1977maximum}, but meanwhile we are assuming that this information is available at the end of each round.

Given the structure of the model; i.e., the variables in it and the directed edges, the joint distribution of those variables can be expressed as a product of the form $P(X_j | Pa(X_j))$ where $Pa(X_j)$ are the parents of $X_j$ in the underlying DAG in $\mathcal{G}$. Since these distributions fully characterize the model, the decision maker will have beliefs over each one of these parameters. Notice that each of these parameters is itself a distribution of length equal to the number of possible values of the variable which is being conditioned, call the maximum number of possible values $k$ . 

A distribution suitable to modelling discrete probability vectors is the $k$-dimensional Dirichlet distribution, whose support is the set of probability vectors of length $k$ \cite{hjort2010bayesian}. The $k$ dimensional Dirichlet distribution has a density $f$ with respect to the Lebesgue measure given by

\[ f(x_1,...,x_k | \alpha_i,...,\alpha_k)=\frac{1}{B(\alpha)}  \prod_{i=1}^k x_i^{\alpha_i-1},\]

where $(x_1,...,x_k)$ are such that $\sum_{i=1}^k x_i =1$ and $\alpha=(\alpha_1,...,\alpha_k)$. The Dirichlet distribution is useful since it is conjugate for itself \cite{bernardo2000bayesian}.

In this way, the decision maker will have beliefs about the CPT's in the form of parameters of several Dirichlet distributions. Using the agent's current beliefs, a causal graphical model can be specified. Using this fully specified (structure + parameters) as a true model, the decision maker will make her choice as in Case 1. When the decision maker observes the value of the target variable, she will update the parameters that specify her beliefs.

Previously we argued that the agent's beliefs were going to be \textit{distributions} over a suitable space, but what is going to be updated are the parameters of such distributions. Namely, the $\alpha$ corresponding to the Dirichlet random variable assigned to each CPT.

For the belief updating, given a new data point, two cases must be considered:
\begin{itemize}
\item The variable to update has no parents.
\item The variable to update has parents.
\end{itemize}

In the first case, if a prior Dirichlet($\alpha$) is used, then the posterior is given by
\[ \textrm{Dirichlet}(\alpha + c) \]
where $c$ is a vector of the number of occurrences of that observed data point. 

For the second case, we must consider both the occurrences of that data point as well as the parents for each of the variables. Following \cite{barber2012bayesian} we denote as $\theta_i(X,j)$ the number of times the event $\{X=i | Pa(X)=j\}$ is observed. In this case, if the prior of $X_i$ conditioned on its parents having the value $j$ is given by a a Dirichlet($\alpha$), then the posterior for the variable $X_i$ given an observed data point is given by 
\[ \prod_j \textrm{Dirichlet}(\alpha + \theta_i(\cdot,j)). \]

\subsection{The model is not known}
The causal model were fully unknown, the decision maker will have to deal with the problem using only any previous knowledge and her own intuitions. Again, any previous knowledge and considerations will be expressed as \textit{beliefs} about the uncertainties in the environment, which will take the form of a probability distributions over a suitable space. 

As in the previous case, we consider a repeated game where the base game consists of Nature assigning a random state of the environment and responding to the agents choices with the effects that were caused by her decisions. In this game, as well as in the previous one, the decision maker will attempt to learn by updating, and using, beliefs in a suitable way. 

The most notable difference with the previous case is that the \textit{structure} of the model is also to be learned in such a way that both the structure and parameters converge to the true model in the limit. In the previous case the decision maker knew the form of the Conditional Probability Tables (CPT) involved in any calculation. In this case, she doesn't know the structure of the DAG so which CPT's are involved is unknown.

If the decision maker knew which variables appear in the true model that governs the environment, even though she didn't know how they are connected, she could use a \textit{Dirichlet Process} to generate Dirichlet distributions and generate causal graphical models the same way as in Case 2 and updating the parameters of the process using the observed information. The Dirichlet Process\footnote{with parameters $M,G_0$}, which was introduced by \cite{ferguson1973bayesian}, is random measure defined over a space $S$ such that for each partition $B_1,...,B_k$ the vector $(G(B_1),...,G(B_k))$ follows a Dirichlet distribution \cite{hjort2010bayesian}, \cite{muller2016bayesian}, \cite{ghosal2017fundamentals}. 

Belief updating using causal information when the decision maker doesn't know the structure of the model nor its parameters is yet to be studied and left as future work. 

\section{Test scenario}
We consider the following hypothetical example where the proposed method will be tested.

Consider a patient who arrives at a hospital who can either have disease $A$ or disease $B$. The doctor can either give him some pill or send him into surgery.  Both treatments entail risks and whether the treatment cures the patient or not depends on which disease it had originally. The doctor could be facing a mutation from a known disease, so she has some knowledge about what could happen if a treatment is given to the patient. Using her previous knowledge as a true model, she can choose a treatment and observe the outcome from which she will learn about this disease, so she could make a better decision the next time a similar patient arrives.

\begin{figure}[ht]
\vskip 0.2in
\begin{center}
\centerline{\includegraphics[width=\columnwidth]{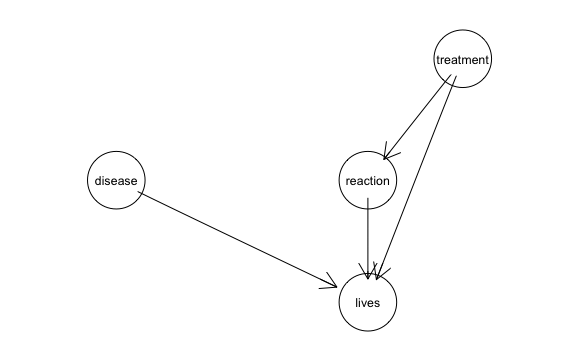}}
\caption{Causal graphical model for the test scenario: the target variable \textit{Lives} is causally influenced by the disease the patient has, the treatment assigned and the survival to the secondary effects of treatment.}
\label{causal_model}
\end{center}
\vskip -0.2in
\end{figure}

The causal model that governs this situation is shown in Figure \ref{causal_model}. The parameters for this model were fixed intuitively in such a way that each treatment is effective for only one disease, but the most effective treatment is riskier.

The variables in the model are: 
\begin{itemize}
\item \textbf{Disease:} Either $A$ or $B$.
\item \textbf{Treatment:} Either pill or surgery.
\item \textbf{Reaction:} Either dying or surviving.
\item \textbf{Lives:} Either living or dying.
\end{itemize}

The variables are causally related as shown in Figure \ref{causal_model}.

The variable \textit{Lives} is the \textit{target variable} and, in this example, the only variable that can be intervened upon is the variable \textit{Treatment}. The decision maker prefers an outcome in which the patient lives.

In this scenario, Nature's move will consist in randomly assigning a disease to the patient. Then, the medic will asign a treatment using his current beliefs about the disease and the possible outcomes. The decision nodes for this play of the medic form an \textit{information set} because the medic doesn't know how she arrived there since she doesn't know what disease did Nature assign. Finally, Nature will sample the consequence of the treatment from the causal model and the medic will observe the outcome.

For this test scenario whose causal graphical model is shown in Figure \ref{causal_model}, we see  by applying the Pearl's do-calculus that the interventional distribution $P_{do(Tr)}(Y)$ is given by
\[ P(Y | do(Tr))=P(Y | D, Tr, R)P(R | Tr) P(D). \]

In fact, from the structure of the model, which is shown in Figure \ref{causal_model}, we see that the involved probabilities in any calculations are:
\[ P(\textrm{disease}), P(\textrm{treatment}), P(\textrm{reaction} | \textrm{treatment}), \]
and
\[P(\textrm{lives} | \textrm{disease, treatment, reaction}). \]

We can also see that the joint distribution for all of the variables can be expressed as
\[ P(Y | D, Tr, R)P(R | Tr) P(D)P(Tr). \]
This expression will be useful when specifying beliefs about the model as Dirichlet distributions.
\section{Experiment}
As proof of concept we implemented  Case 1 and Case 2 for the test scenario and compared it with an agent performing Q-learning \cite{watkins1992q} and an agent choosing her actions at random. 

For the implementation, we defined a \textit{true} causal model as the one shown in Figure \ref{causal_model} using the library Pgmpy \cite{ankan2015pgmpy}. Then, we defined an agent which has beliefs for each of the CPT that appear in the factorization for the model and randomly assigned values for the $\alpha$ parameters for each one of the Dirichlet distributions as mentioned before. 

This agent will find the action that maximizes her desired value for the target variable using do-calculus. The action thus selected will be used as \textit{evidence} in the true causal model and a MAP inference will be used to simulate the most likely outcome given this action. The target variable will output a $1$ value if the patient lives. The value of the target variable will be the reward of each round.

\subsection{Case 1: The causal model is completely known}
If the causal model is completely known to the decision maker, then in one step she can obtain the probability for her desired value of the target variable, which in this case is the value corresponding to the outcome in which the patient lives at the end. Using this probability, she can choose which treatment to assign. Since this action maximizes the probability of the occurence desired value, it maximizes the expected utility, and it is also a \textit{best response} to the player Nature.

\subsection{Case 2: Only the structure is known}
From the expression of the joint probability we notice that we need to specify a distribution over each one of these distributions, which will be each one a Dirichlet distribution.

We begin with a random assignation of the $\alpha$ parameter for each of the distributions considered. We use Dirichlet distribution for each of the conditional probability tables that appear in the factorization of the joint probability for the graph of $\mathcal{G}$. Since each of the variables in the model is binary, then the product of these Dirichlets is again Dirichlet.

With this parameters, the decision maker forms a causal model and chooses the action that maximizes the probability of the desired value for target variable as in Case 1. With this action chosen, we simulate an outcome from the causal graphical model using the chosen action as an intervention. This evidence is used to update the parameters, which then will be used to generate a new causal model, and so on.

We show the results of the experiment, where we compare the performance obtained by the causal agent, a \textit{random agent} who selects his actions at random, and an agent performing Q-learning. We show the average reward obtained by the agent over $20, 50, 100$ and $200$ rounds.

\begin{figure}[ht]
\vskip 0.2in
\begin{center}
\centerline{\includegraphics[width=\columnwidth]{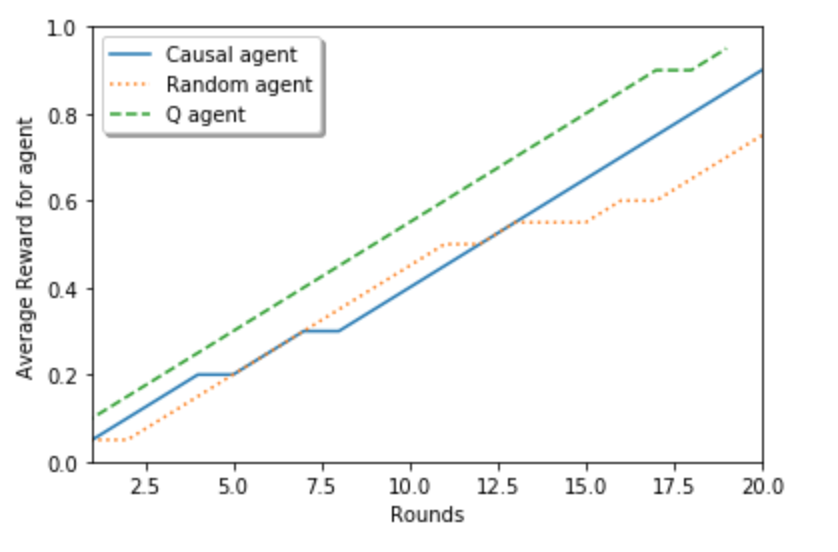}}
\caption{Average reward obtained in each round for each agent}
\label{20_rounds}
\end{center}
\vskip -0.2in
\end{figure}

In Figure \ref{20_rounds} we observe the average rewards for each agent in 20 rounds of decision making. Here we notice that Q-learning outperforms our algorithm, which has a similar performance as the random choosing procedure until round 11.

\begin{figure}[ht]
\vskip 0.2in
\begin{center}
\centerline{\includegraphics[width=\columnwidth]{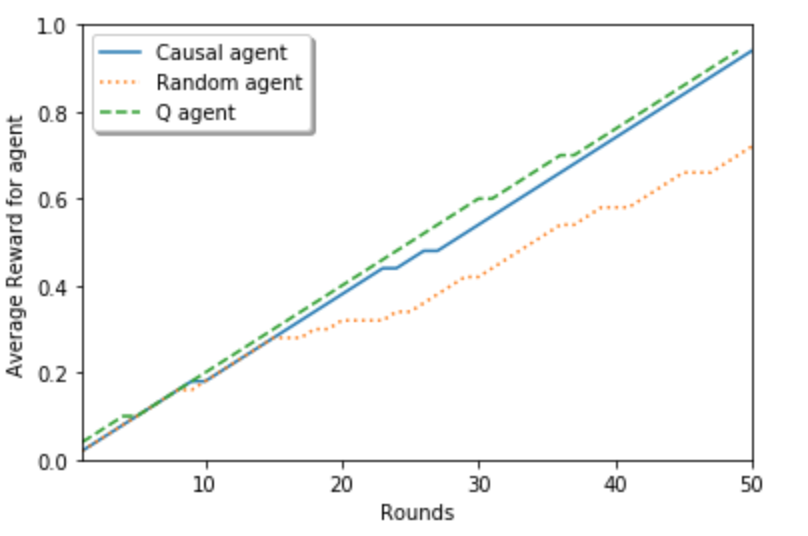}}
\caption{Average reward obtained in each round for each agent}
\label{50_rounds}
\end{center}
\vskip -0.2in
\end{figure}

In Figure \ref{50_rounds} we observe the average rewards for each agent in 50 rounds of decision making. Our algorithm follow closely the Q-learning agent and outperform the random agent.

\begin{figure}[ht]
\vskip 0.2in
\begin{center}
\centerline{\includegraphics[width=\columnwidth]{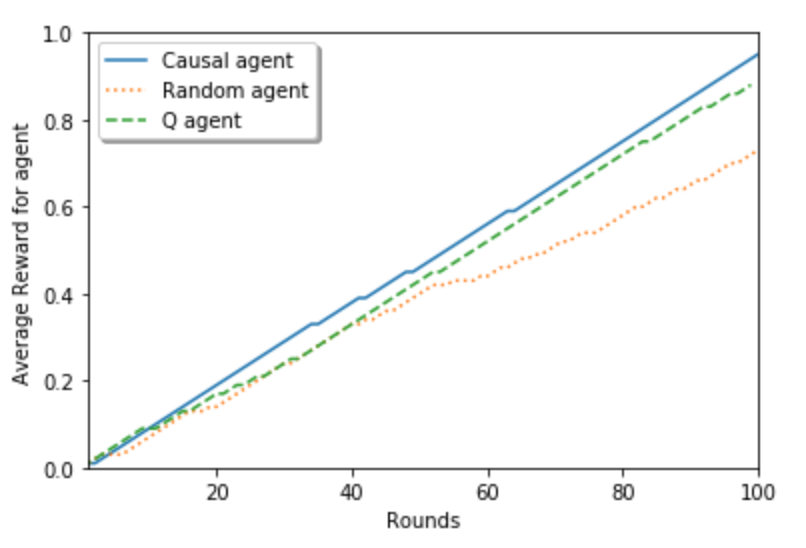}}
\caption{Average reward obtained in each round for each agent}
\label{100_rounds}
\end{center}
\vskip -0.2in
\end{figure}

In Figure \ref{100_rounds} we observe the average reward obtained by the three agents in 100 rounds, where our algorithm slightly outperforms Q-Learning.

\begin{figure}[ht]
\vskip 0.2in
\begin{center}
\centerline{\includegraphics[width=\columnwidth]{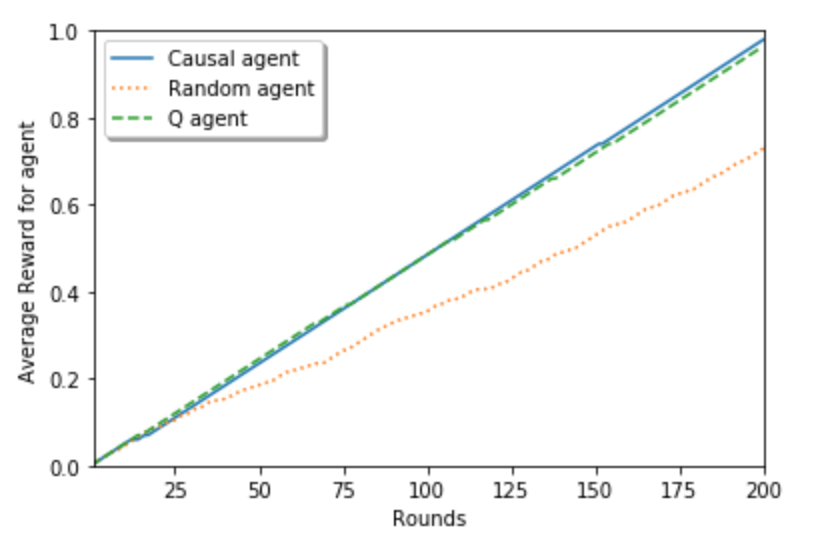}}
\caption{Average reward obtained in each round for each agent}
\label{200_rounds}
\end{center}
\vskip -0.2in
\end{figure}

In Figure \ref{200_rounds} we observe the average reward obtained by the three agents in 200 rounds. The average reward obtained is very similar for Q-learning and our algorithm.

We see that our method obtains a very similar reward as the classic Q-learning algorithm for a larger number of rounds, where the random agent is outperformed. Even though our model has a similar performance to a classic learning algorithm it learns a causal model of the environment which allows it to explain why an agent chose her actions as well as allowing conterfactual reasoning.

\section{Conclusion and Future Work}
In this work we have proposed a way to make decisions in uncertain environments which are known to be governed by causal mechanisms. The proposed decision making procedure attempts to resemble how human beings act when causal relations are present. Human beings are known to use, and modify, causal knowledge when making decisions. 

This ideas motivated us to study how an autonomous agent could learn from her environment when it has a particular structure, in this case being causal relations what gives the environment a certain structure. 

We assumed here that the decision maker is aware of the causal nature of the environment, but lacks information about its specific parameters which are to be learned by interaction. It is reasonable to assume that the variables involved are known, since in many situations we are aware of what we are intervening upon and what do we expect it to affect. 

The experimental results show that considering causal structure in a decision making process yields a good performance when compared to non-causal classic algorithms, but our model has the extra feature of learning a causal model of the environment which could be exported for problems involving similar scenarios.

The problem of discovering the variables itself and the connections between them is far more general and it is left as future work.

\section{Acknowledgments}
We are grateful to the National Institute of Astrophysics Optics and Electronics (INAOE) and to Mrs. Graciela Soto for their generous funding in order to attend ICML 2018. 

\bibliography{Bibliografia.bib}
\bibliographystyle{icml2018}

\end{document}